\documentclass[journal]{IEEEtran}
\usepackage{cite}
\usepackage[pdftex]{graphicx}
\usepackage{amsmath}
\usepackage{array}
\usepackage{amssymb}
\usepackage{xcolor}
\usepackage{booktabs}
\usepackage{multirow}
\usepackage{bm}
\usepackage{subcaption}
\usepackage[autolanguage]{numprint}
\PassOptionsToPackage{hyphens}{url}
\usepackage{hyperref}

\makeatletter
\renewcommand\p@paragraph{\thesubsection.\expandafter\@gobble}
\makeatother

\begin{document}

\title{Power Flow Balancing with Decentralized Graph Neural Networks}

\author{Jonas Berg~Hansen,
        Stian Normann~Anfinsen,
        and~Filippo Maria~Bianchi
\thanks{Correspondence: \texttt{jonas.b.hansen@uit.no}}
\thanks{J. B. Hansen and F. M. Bianchi are with Department of Mathematics and Statistics, UiT The Arctic University of Norway}
\thanks{S. N. Anfinsen and F. M. Bianchi are with NORCE Norwegian Research Centre AS}
\thanks{S.~N.~Anfinsen is with Department of Physics and Technology, UiT The Arctic University of Norway}
\thanks{This project was supported in part by Equinor under the Academia Agreement with UiT The Arctic University of Norway.}
}

\maketitle

\begin{abstract}
We propose an end-to-end framework based on a Graph Neural Network (GNN) to balance the power flows in energy grids.
The balancing is framed as a supervised vertex regression task, where the GNN is trained to predict the current and power injections at each grid branch that yield a power flow balance.
By representing the power grid as a line graph with branches as vertices, we can train a GNN that is accurate and robust to changes in topology.
In addition, by using specialized GNN layers, we are able to build a very deep architecture that accounts for large neighborhoods on the graph, while implementing only localized operations.
We perform three different experiments to evaluate: i) the benefits of using localized rather than global operations and the tendency of deep GNN models to oversmooth the quantities on the nodes; ii) the resilience to perturbations in the graph topology; and iii) the capability to train the model simultaneously on multiple grid topologies and the consequential improvement in generalization to new, unseen grids.
The proposed framework is efficient and, compared to other solvers based on deep learning, is robust to perturbations not only to the physical quantities on the grid components, but also to the topology.
\end{abstract}

\IEEEpeerreviewmaketitle

\section{Introduction}

\IEEEPARstart{P}{ower} flow analysis is a fundamental part of the operation, planning, maintenance, and control of power systems~\cite{salam2020fundamentals}. To ensure adequate system reliability for day-to-day operation,  the power flow must be distributed such that the demand is met without violating the physical constraints of the electrical components of the energy grid. In addition, the power system should be reasonably resilient to outages and disconnections. To perform the computations necessary to find this balanced state of operation, a power solver needs information about how the electrical nodes in the power grid are connected, as well as the physical properties of the infrastructure (e.g., resistances of the transmission lines).

Traditionally, the solution to the power flow (PF) problem is found using numerical optimizers, such as the Newton-Raphson (NR) method~\cite{8634940}. These methods are computationally expensive for large systems, the convergence to the global optimum is in general not guaranteed, and a new optimization must be performed for each small modification of the grid configuration.   

Recently, several deep learning methodologies have been proposed to solve a wide range of energy analytics problems encountered in power systems~\cite{ozcanli2020deep, khodayar2020deep, zhao2021chapter}. Of particular interest for this work are the approaches that compute fast approximations of the power flow solution while being more accurate than traditional approximations, such as the linear DC power flow (DCPF) method. Most of the deep learning approaches proposed for solving the PF problem, as well as the optimal power flow (OPF) problem, utilize models based on multi-layer perceptrons (MLPs)~\cite{paucar2002artificial,donnot2018fast, singh2021learning, zamzam2020smartgridcomm}. MLPs are easy to implement and they generalize well to different system values (e.g., loads on the buses). However, they struggle with changes in the underlying system topology. 
In addition, MLPs have a higher tendency to overfit the training data compared to other neural networks, which exploit inductive biases to reduce the number of trainable parameters.

A few recent works proposed to represent power systems as graphs and used graph neural networks (GNNs) to solve power flow related tasks~\cite{donon2019graph1, donon2020graph2, owerko2020optimal, bolz2019power, kim2019graph, wang2020probabilistic, nauck2021predicting}. 
GNNs are neural network models that directly exploit the topology of the graph to implement localized computations, which are independent from the global structure of power systems.
Therefore, GNNs can, in principle, be applied to grids with different sizes and topologies. 
The latter is one of the most appealing features of GNN models compared to MLPs, as it allows to quickly compute a solution after the grid topology is modified, e.g.\ when transmission lines are disconnected due to contingencies or planned rerouting operations. 

However, most of the GNN-based approaches found in the existing literature use architectural components that depend on the global structure, which essentially locks the models to a specific topology~\cite{owerko2020optimal, bolz2019power, kim2019graph, wang2020probabilistic}. 
Like the MLP, these models are still effective when dealing with a single grid topology, but are unable to handle topological variations in the configuration of power systems. 
There are a few works that adopt GNNs able to process multiple topologies. 
However, these architectures are either unable to process transmission line attributes, such as resistance or reactance, which are important to model the physics of the power flow problem~\cite{donon2019graph1}, or they use the GNN to perform a single operation within a larger pipeline that includes global operations to update the state of the buses~\cite{donon2020graph2}.
\newline

\textbf{Contributions.}
We propose a fully decentralized GNN architecture and implement an end-to-end data-driven approach to balance the power flows.
The optimization is framed as a supervised vertex regression task, which aims to predict the electrical quantities of interest at each branch.
By using only local operations, our GNN can handle perturbations in the graph topology and, in principle, can even be transferred to grids with different topologies.

Like the DC approximation, the proposed GNN model has a computational complexity that scales linearly (both in time and space) with the number of nodes.
On the other hand, the GNN solution is generally much closer than the DC approximation to the solution found by NR.
In addition, we use a particular graph representation and a GNN architecture that offers flexibility and performance at the same time. 
This allows us to obtain competitive results on a larger variety of tasks, compared to other deep learning architectures.

To effectively implement such a model, we have addressed two challenges:
\begin{itemize}
    \item In power flow balancing, distant vertices on the graph should be able to communicate with each other. 
    To accomplish this in the absence of centralized operations, a very deep GNN with a large receptive field is required.
    Since deep GNN architectures are often affected by oversmoothing (the output of adjacent vertices tends to become the same), we use convolutional ARMA layers~\cite{bianchi2021graph}, which are designed to implement a large receptive field.
    \item 
    Buses with generators must be treated differently when computing the PF solution, which complicates the GNN training.
    In addition, most GNN models can handle only attributes on the vertices (buses) but not on the edges (branches).
    As a solution, we introduce a \textit{line graph} representation of the power grid, which: i) allows the GNN to treat all vertices equally; and ii) allows to represent both bus and branch features as vertex attributes.
\end{itemize}

We perform three experiments to investigate the potential of a decentralized GNN for power flow balancing.
\begin{enumerate}
    \item We compare the proposed GNN with an MLP to highlight the benefits of using localized rather than global operations. We also analyze the problem of oversmoothing that GNNs encounter in power flow balancing and we show how it can be mitigated by using specialized ARMA layers to implement very deep architectures.
    \item We consider a setting where the topology changes as a consequence of line removal in the grid and we show the advantage of using a decentralized architecture.
    \item We perform training on multiple power systems with completely different topologies to investigate if the generalization capability on new, unseen grids improves.
\end{enumerate}

\section{Theoretical Background}

\subsection{Notation}
\label{sec:notation}
A power grid consists of a set of buses and branches that connect them. Each of these core components are characterized by a set of physical quantities, which influence the power flow solution.
Those are referred to as \textit{features} and are described in the following, along with their measurement units (in brackets). Note that most quantities are given in per unit (p.u.), which are derived from the base power and voltage levels of the grid. A reasoning behind this choice of units is given when presenting the data acquisition in Section~\ref{sec: datasets_and_algorithms}.

\begin{enumerate}
    \item Bus features:
    \begin{itemize}
        \item $|V|$: voltage magnitude [p.u.]
        \item $\theta$: voltage angle [radians]
        \item $P_d$: active power demand [p.u.]
        \item $Q_d$: reactive power demand [p.u.]
        \item $G_s$: conductance of connected shunt element [p.u.]
        \item $B_s$: susceptance of connected shunt element [p.u.]
        \item $P_g$: active power output of connected generator [p.u.]
        \item $Q_g$: reactive power output of connected generator [p.u.]
    \end{itemize}
    \item Branch features:
    \begin{itemize}
        \item $r/x$: resistance/reactance [p.u.]
        \item $b$: total line charging susceptance [p.u.]
        \item $\tau$: transformer off nominal turns ratio
        \item $\theta_\text{shift}$: transformer phase shift angle [radians]
    \end{itemize}
\end{enumerate}

\subsection{The Power Flow Problem}

A distinction must be made between power flow (PF) -- also called AC power flow (ACPF) -- and optimal power flow (OPF) problems; while the former is solving a system of equations, the latter minimizes a cost function subject to constraints that include the PF equations. This paper focuses on PF.

The steady-state solution to the PF problem for a system of $N_b$ buses is found by solving a set of nonlinear equations given by Kirchhoff's laws. In polar form, the power balance equations are~\cite{milano2010power}:
\begin{align}
    P_i &= \sum_{k=1}^{N_b} |V_i||V_k|(G_{ik}\cos\theta_{ik} + B_{ik}\sin{\theta_{ik}}), \label{eq: act_power_balance}\\
     Q_i &= \sum_{k=1}^{N_b} |V_i||V_k|(G_{ik}\sin\theta_{ik} - B_{ik}\cos{\theta_{ik}}) \label{eq: react_power_balance},
\end{align}
where $P_i$ and $Q_i$ are the net active and reactive power injections at bus $i$, $\theta_{ik}$ is the voltage phase angle difference between bus $i$ and $k$, and $B_{ik}$ and $G_{ik}$ are respectively the real and imaginary parts of the $(i,k)$-th element in the bus admittance matrix. The shunt conductance and susceptance, along with all of the aforementioned branch attributes, are used to construct this admittance matrix.

Typically, the aim of power flow balancing is to find the complex voltages on the buses that satisfy the balance equations in \eqref{eq: act_power_balance} and \eqref{eq: react_power_balance}. 
To obtain the same number of equations and variables, buses are split into different categories where certain variables are fixed. 
Firstly, a single bus with a generator is selected to act as the slack or reference bus, for which $|V_i|$ and $\theta_i$ are fixed. The voltage phase angles of the other buses are expressed as differences from the slack. The other buses with generators are called PV buses, and for these $|V_i|$ and $P_i$ are fixed. Finally, the remaining buses are called PQ buses, where $P_i$ and $Q_i$ are known.

\subsection{Multi-Layer Perceptrons}
An MLP consists of multiple layers of computational (or processing) units that perform an affine transformation followed by a nonlinearity~\cite{goodfellow2016deep}.
Formally, an MLP with $L$ hidden layers implements the following operations
\begin{align}
    \bm{h}^{(0)} &= \bm{x},\\
    \bm{h}^{(l)} &= \sigma\left(\mathbf{W}^{(l)}\bm{h}^{(l-1)} + \bm{b}^{(l)}\right),\hspace{0.3cm}l=1,2,...,L, \label{eq: mlp_mid_layer}\\
    \bm{y} &= \sigma\left(\mathbf{W}^{(L+1)}\bm{h}^{(L)} + \bm{b}^{(L+1)}\right),\label{eq: mlp_final_layer}
\end{align}
where $\bm{x}\in\mathbb{R}^{F_\text{in}\times 1}$ is the input, $\bm{h}^{(l)}\in\mathbb{R}^{F_\text{l}\times 1}$ is the hidden state at the $l$-th layer, $\bm{W}^{(l)}\in\mathbb{R}^{F_l\times F_{l-1}}$ is a trainable weight matrix, $\bm{b}^{(l)}\in\mathbb{R}^{F_l\times 1}$ is a trainable bias vector, $\sigma$ is an activation function, and $\bm{y}\in\mathbb{R}^{F_\text{out}\times 1}$ is the MLP output. $F_l$ denotes the number of processing units in the $l$-th layer. When used as intermediate or final layers in other artificial neural network models, the operations performed by \eqref{eq: mlp_mid_layer}-\eqref{eq: mlp_final_layer} are often referred to as dense or fully connected (FC) layers.

The dimensions of the weight matrices and bias vectors in the input and output layers depend on the dimension of $\bm{x}$ and $\bm{y}$.
This presents a problem when the MLP is applied to data where the number of variables differ across samples.
Such problems can be addressed by workarounds such as using a super-set of variables and then mask the unused variables in each sample.
This, however, poses issues in a PF setting because the MLP cannot learn relationships between model variables and the actual electrical elements if they change from sample to sample. Also, it is not possible to process test samples with more than the maximum number of variables allocated in training.
In addition, when most samples have fewer variables than the maximum allowed, some of the model capacity remains unused.
Finally, as the number of trainable parameters increases with the number of variables, MLPs used to process large grids must have a very high capacity and can more easily memorize specific patterns in training data at the expense of generalization to new data. This problem is referred to as overfitting~\cite{goodfellow2016deep}.

\subsection{Graph Neural Networks}
\label{sec:GNNs}
Power systems can be naturally represented as graphs, due to their non-Euclidean structure.
A graph with $N$ vertices, each one associated with a feature vector of size $F_\text{in}$, can be represented by the tuple $\{ \mathbf{A}, \mathbf{X} \}$, where $\mathbf{A}\in\mathbb{R}^{N \times N}$ is the adjacency matrix and $\mathbf{X}\in\mathbb{R}^{N \times F_\text{in}}$ contains the vertex features.
GNNs are a class of neural networks specifically designed to process data represented as graphs~\cite{bacciu2020gentle}.
GNNs include layers implementing localized operations, called \textit{message passing} (MP), that are independent from the global structure of the graph.
A GNN without layers that perform global operations 
is completely decentralized and can process graphs with different topologies. 

The MP operations in a GNN can be performed in several different ways.
A popular MP layer is the one from the Graph Convolutional Network (GCN)~\cite{kipf2017semi}, which updates the node features as
\begin{equation}
    \label{eq:gcn}
    \bar{\mathbf{X}}^{(l)} = \sigma\left(\hat{\mathbf{A}} \bar{\mathbf{X}}^{(l-1)} \mathbf{W} \right)
\end{equation}
where $\bar{\mathbf{X}}^{(l)}$ are the vertex features at layer $l$ (note that $\bar{\mathbf{X}}^{(0)} = \mathbf{X}$), $\mathbf{W}\in\mathbb{R}^{F_\text{in} \times F_\text{out}}$ is a matrix of trainable parameters, and 
$\hat{\mathbf{A}}$ is the symmetrically normalized version of $\mathbf{A} + \mathbf{I}$, given by
\begin{align}
    \hat{\mathbf{A}} &= \hat{\mathbf{D}}^{-1/2} (\mathbf{A} + \mathbf{I}) \hat{\mathbf{D}}^{-1/2},
\end{align}
where $\hat{\mathbf{D}}$ is the degree matrix of $(\mathbf{A} + \mathbf{I})$.

\begin{figure}[t!]
    \centering
    \includegraphics[width=\columnwidth]{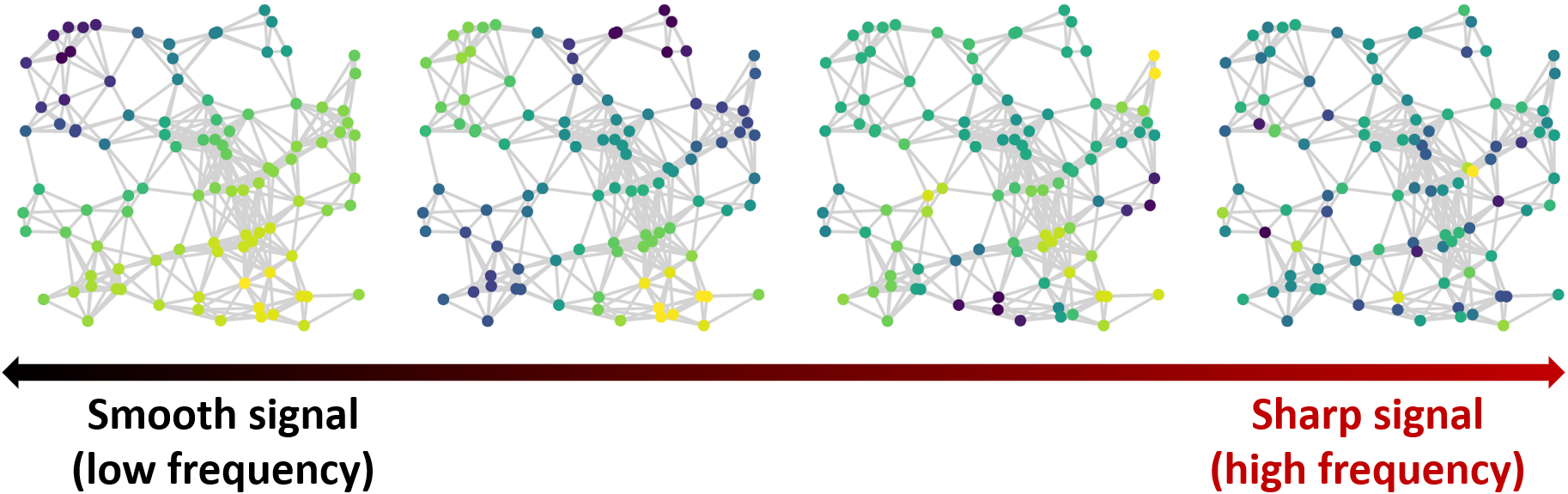}
    \caption{The graph signal becomes sharper going from left to right. Smooth graph signals assign similar values to neighboring nodes, while a sharp graph signal can assign different values to neighboring nodes. Most GNN layers perform a smoothing operation and, thus, by stacking many of them it is possible to obtain in output only very smooth graph signals.}
    \label{fig:oversmoothing}
\end{figure}

To optimize the power flow, it is necessary to reach higher order neighborhoods, since the value of a generator can be influenced by the load of a distant bus in the grid. 
To combine nodes from distant regions in the graph, it is necessary to stack many MP operations. If these operations are implemented by layers that smooths graph signals \cite{kipf2017semi, velivckovic2018graph, hamilton2017inductive}, the repeated application of MP leads to \textit{oversmoothing}, i.e., the vertex representations become uniform on the whole graph~\cite{li2018deeper}. 
This prevents the GNN from producing a dissimilar output at vertices that are very close on the graph. 
The difference between smooth and sharp graph signals is illustrated in Fig.~\ref{fig:oversmoothing}. 

It is important for the power flow problem that a GNN can generate sharp signals, since neighboring grid buses might assume very different physical quantities.
As a solution, we implement a very deep GNN composed of convolutional ARMA layers \cite{bianchi2021graph}. The ARMA layer is more flexible and less prone to oversmooth the vertex features compared to other MP layers. 
Each ARMA layer consists of $K$ stacks, each one performing $T$ propagation steps, which we will denote as ARMA$_K^T$. The output of each stack is computed as
\begin{align}
    \bar{\mathbf{X}}^{(1)}_k &= \sigma\left(\Tilde{\mathbf{A}} \mathbf{X} \mathbf{W}_k^{(0)} + \mathbf{XV}_k^{(0)}\right), \label{eq: armait_1}\\
    \bar{\mathbf{X}}^{(t)}_k &= \sigma\left(\Tilde{\mathbf{A}} \bar{\mathbf{X}}^{(t-1)}_k \mathbf{W}_k + \mathbf{XV}_k\right),\hspace{0.25cm} t=2,3,...,T, 
\label{eq: armait_t}
\end{align}
where $\bar{\mathbf{X}}_k^{(t)}\in \mathbb{R}^{N\times F_\text{out}}$ are node features at step $t$, $\sigma$ is a nonlinear activation function, and $\mathbf{W}_k^{(0)}$, $\mathbf{V}_k^{(0)}$, $\mathbf{V}_k$ (all $\in\mathbb{R}^{F_\text{in} \times F_\text{out}}$) and $\mathbf{W}_k\in\mathbb{R}^{F_\text{out} \times F_\text{out}}$  are trainable weight matrices.
$\Tilde{\mathbf{A}}$ is the symmetrically normalized version of the adjacency matrix $\mathbf{A}$, i.e., $\Tilde{\mathbf{A}} = \mathbf{D}^{-1/2} \mathbf{A} \mathbf{D}^{-1/2}$.

The final output for the ARMA$_K^T$ layer is computed as an average of the output of all stacks:

\begin{equation}
    \bar{\mathbf{X}} = \frac{1}{K}\sum_{k=1}^K \bar{\mathbf{X}}_k^{(T)}
    \label{eq: arma_average}
\end{equation}

\section{Methodology}
\label{sec: method}

\subsection{Power grid representation}

When representing a power grid as a graph it is natural to let buses act as graph vertices and branches as graph edges. However, for a fully data-driven GNN architecture relying solely on localized operations, such a representation raises two main issues.

Firstly, in a bus-oriented graph, the voltage magnitude associated to specific vertices is fixed and should remain unchanged in the output computed by the GNN.
However, given the decentralized nature of a GNN, which processes all vertices with the same localized operations, treating certain vertices categorically different is problematic and the output of the GNN will need to be adjusted for these vertices.

The more challenging problem is how to account for features from both buses and branches. 
In a setting where the model encounters only a single, static grid, it is sufficient to only include variables from buses that vary from sample to sample.
Indeed, quantities such as line resistances can be factored out, as they are common to every sample.
On the other hand, we are interested in processing samples that represent modifications of a given topology or even completely different power systems.
In this latter case, to learn the interactions between all the physical quantities in the power system, the model needs access to all relevant bus and branch features.
There exist few GNN models in the literature that are designed to process edge features/labels.
For instance, an edge-conditioned convolutional layer~\cite{Simonovsky_2017_CVPR} can be used to embed the information of branch features into the vertex representations.
However, these layers have a significantly higher computational cost than those that consider only vertex features. Also, since the relationship between vertices change as their features are propagated, a layer that accounts for original edge information is mostly effective if used in the beginning of the GNN architecture. 
However, if the GNN is very deep, like the one needed to solve the PF problem of interest, the contribution of the first layer to the computation of the output is weak and difficult to control. 

To address these issues, we propose to represent the branches as graph vertices, and to predict the complex current and power injections on the branches, rather than the voltage on the buses. 
These power and current quantities can be obtained from the complex bus voltages and can be used to express an equivalent power flow solution. 
Since none of these quantities are fixed beforehand, it is not necessary to constrain parts of the GNN output to assume specific values. 
Finally, since each branch is connected to exactly two buses, features from both branches and buses can be placed on the graph vertices; this allows us to use GNN architectures that only consider vertex features.

\begin{figure}[!t]
    \centering
    \includegraphics[width=0.35\textwidth]{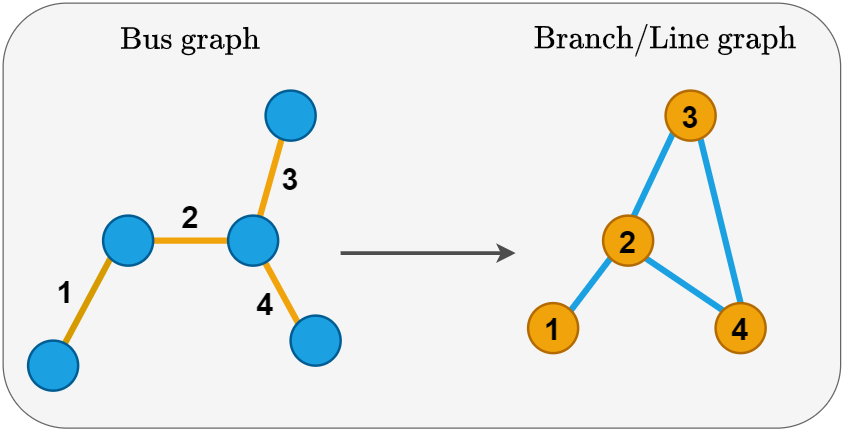}
    \caption{Transition from the bus graph ($\mathcal{G}$) to the branch graph (line graph of $\mathcal{G}$).}
    \label{fig:bus_to_branch_graph}
\end{figure}
The shift in perspective from buses to branches can be seen as a conversion to a \textit{line graph} representation \cite{harary1960some} of the bus graph. 
Given a graph $\mathcal{G}$, its line graph $L(\mathcal{G})$ is a graph where each vertex represents an edge of $\mathcal{G}$. Two vertices of $L(\mathcal{G})$ are adjacent if and only if their corresponding edges in $\mathcal{G}$ are incident, i.e., they share a common endpoint.
This yields an undirected graph, as illustrated in Figure \ref{fig:bus_to_branch_graph}. 

Since each branch has a form of directionality related to the assignment of designated \textit{from} and \textit{to} ends for transformers, a directed graph could be used to model the branches' direction.
However, in a directed graph the peripheral vertices at the end of a chain of unidirectional edges cannot share (or receive, depending on the edges direction) information with the rest of the graph.
This affects the learning process of a GNN, as the graph diffusion is hindered.
Therefore, we have opted to keep the undirected graph representation and the direction is implicitly expressed through the inclusion of bus features from the two endpoints of the branches.

Each vertex in the line graph is associated with the following features
\begin{itemize}
    \item From corresponding branch: $r$, $x$, $b$, $\tau$, $\theta_\text{shift}$
    \item From bus endpoints: $P_d$, $Q_d$, $G_s$, $B_s$, $P_g$, $|V_g|$, $\bm{i}_\text{slack}$
\end{itemize}
where $P_g$ and $|V_g|$ are the active power and voltage magnitude of a connected generator, respectively. Finally, $\bm{i}_\text{slack}$ is a two-dimensional one-hot vector indicating whether the bus at the corresponding endpoint is the slack bus. Naturally, not all buses have generators or shunts, and only select branches are equipped with transformers, so in these cases the corresponding features (such as $|V_g|$) are set to zero to indicate absence.

To summarize, each vertex $i$ in the line graph is associated with a vector $\mathbf{x}_i \in \mathbb{R}^{21}$ of input features: [branch(5); \textit{from}-bus(8); \textit{to}-bus(8)].
By consistently concatenating the features of the \textit{from}-bus before the features of the \textit{to}-bus, the branch direction is implicitly given.

As for the desired PF solution with respect to the branches, injections are made at both ends of a branch, so if power and current are split into their real and imaginary parts, the power flow optimization is framed as a vertex regression task, where the GNN predicts 8 quantities for each branch. 
Specifically, the output features to be approximated by the GNN consist of a vector $\mathbf{y}_i \in \mathbb{R}^8$ containing the following quantities
\begin{itemize}
    \item $P^\text{re}_f, P^\text{im}_f$, $I^\text{re}_f, I^\text{im}_f$, $P^\text{re}_t, P^\text{im}_t$, $I^\text{re}_t, I^\text{im}_t$
\end{itemize}
which are the real and imaginary components of power and current at the two endpoints ($f=\textit{``from''}$ and $t=\textit{``to''}$).

The choice of using power flow and current is particularly suitable for the line graph representation. Firstly, power flow and current are properties that relate more to the branches than the voltages on the buses. Secondly, since the same voltage can appear on multiple vertices of the line graph, one would have to enforce constraints to the output to ensure that such voltage values are the same.

\subsection{GNN Design}
The node aggregation of the GNN model proposed here is done with ARMA$_k^T$ layers. 
As discussed in Section~\ref{sec:GNNs}, by using ARMA layers it is possible to build very deep GNN architectures that are robust to oversmoothing.
In principle, one could use a single ARMA layer with a very high number of iterations $T$ to achieve the desired depth of the receptive field.
However, we found out experimentally that a GNN with more ARMA layers, each one with fewer recursions, performs better.

\begin{figure}[t]
    \centering
    \includegraphics[width = 0.35\textwidth]{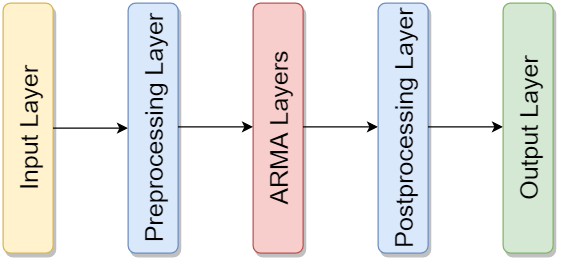}
    \caption{General structure of the GNN model.}
    \label{fig:gnn_architecture}
\end{figure}

Before and after the aggregation stage, the model uses dense processing layers to map the individual node states. The idea here is to allow the model to map the states into a more meaningful representation before the aggregation, and to map the final aggregated states into the output space. This type of configuration is aligned with the study on the architectural design space of GNNs by \cite{you2020design}, which showed that such pre- and post-processing layers could improve performance when combined with different aggregation techniques.
Fig.~\ref{fig:gnn_architecture} shows a schematic depiction of the adopted GNN architecture.

The main advantage of using GNNs compared to NR is the lower time and space complexity. The number of operations performed when evaluating the expression within the activation of either \eqref{eq: armait_1} or \eqref{eq: armait_t} is $(N \times N \times F_\text{in}) + (N \times F_\text{in} \times F_\text{out}) + (N \times F_\text{in} \times F_\text{out}) = (N^2 \times F_\text{in}) + 2(N \times F_\text{in} \times F_\text{out})$. Therefore, an ARMA layer with $K$ stacks of depth $T$ performs $K\times T\times ((N^2 \times F_\text{in}) + 2(N \times F_\text{in} \times F_\text{out}))$ operations and, thus, the time complexity is $\mathcal{O}(N^2)$.
Similarly, a stack of $T$ GCN layers performs $T \times ((N^2 \times F_\text{in}) + (N \times F_\text{in} \times F_\text{out}))$ operations and the complexity is again $\mathcal{O}(N^2)$.
Under the sparsity assumption, the number of non-zero connections in the graph is proportional to $N$ and thus, by using sparse operations, both the space and time complexity of the GNNs reduces from $\mathcal{O}(N^2)$ to $\mathcal{O}(N)$.

\subsection{MLP Design}
MLP models will be used to benchmark the GNN architecture. 
We consider two different configurations of MLPs, one that processes each vertex features vector individually and another that accounts for all vertices at once.

\begin{figure}[ht!]
    \centering
    \includegraphics[width=0.45\textwidth]{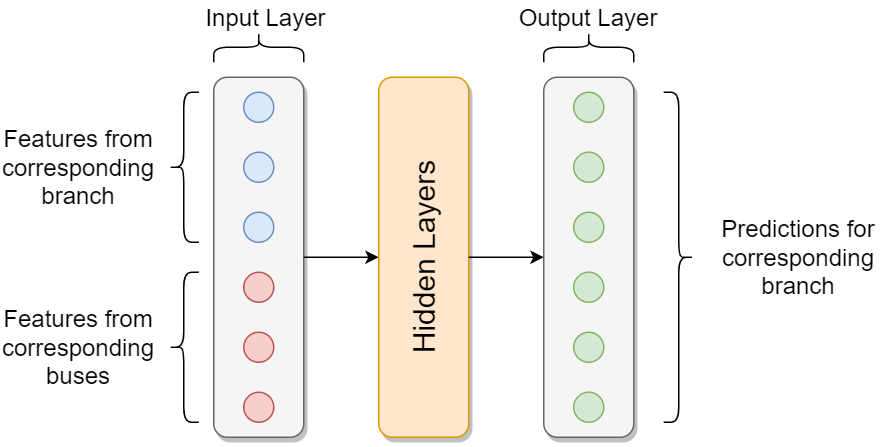}
    \caption{Local MLP. The number of units shown here does not reflect the exact amount used for the experiments.}
    \label{fig:local_mlp_architecture}
\end{figure}

The first configuration, which we call \textit{Local MLP model}, makes predictions for each branch using only its associated input features. In other words, the 8 target values for each branch are predicted using the aforementioned 21 input vertex features. See Fig.~\ref{fig:local_mlp_architecture} for an illustration.
Since the Local MLP is completely agnostic of the underlying topology of the grid, the performance obtained by the Local MLP model can be used to quantify how much the information contained in the input features associated with a single branch and its endpoints can, alone, predict the output values for that branch.

\begin{figure}[ht!]
    \centering
    \includegraphics[width=\columnwidth]{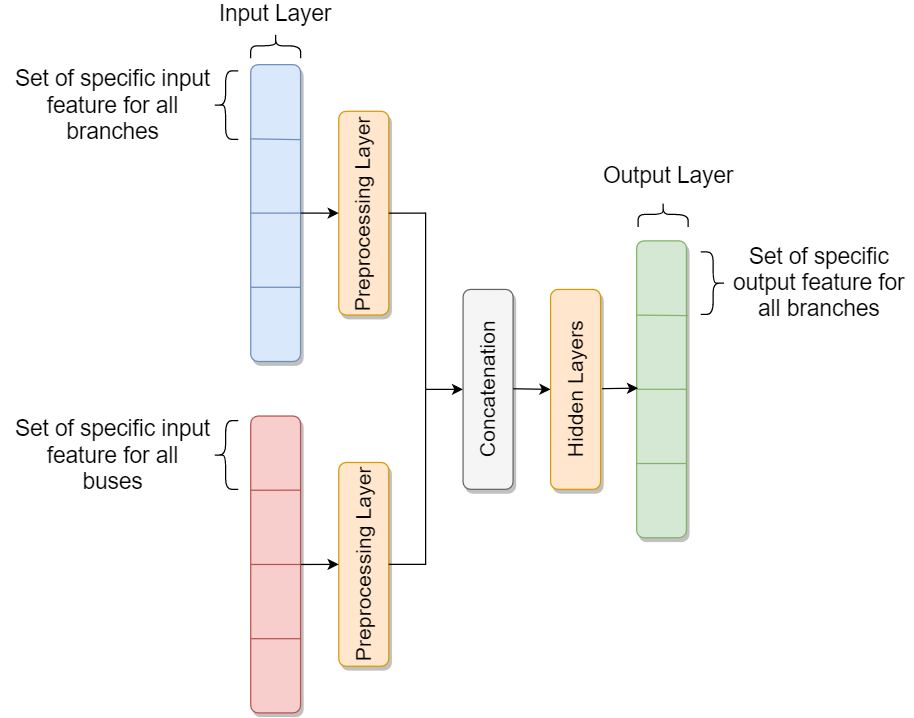}
    \caption{Global MLP.}
    \label{fig:global_mlp_architecture}
\end{figure}

In the second configuration, referred to as the \textit{Global MLP model}, predictions are made for all branches at once using the input features from all buses and branches. To accomplish this, all bus and branch input features are concatenated into two one-dimensional arrays. Moreover, to reduce the number of parameters in the input layer, bus and branch features are first processed by separate layers, and the output of these layers are then concatenated and fed to the hidden layers of the model, as shown in Fig.~\ref{fig:global_mlp_architecture}. At the end, this MLP model predicts all the branch currents and power injections, i.e.\ the number of output units is equal to eight times the number of branches.
Clearly, the number of parameters of the Global MLP depends on the number of vertices and edges in the graph, which makes the model less suitable to handle perturbations on the grid topology.

\section{Related work on power flow balancing}

\paragraph{Newton-Raphson}
The traditional Newton-Raphson (NR) method iteratively solves a set of non-linear equations through a linear approximation based on a first order Taylor polynomial. For the power flow problem this results in a linear system of equations of the form:
\begin{equation}
    \label{eq:newton_raphson}
    \begin{bmatrix}
    \Delta \theta_i \\
    \Delta |V_i|
    \end{bmatrix} = 
    -J^{-1}
    \begin{bmatrix}
    \Delta P_i \\ \Delta Q_i
    \end{bmatrix},  
\end{equation}
where $\Delta P_i$ and $\Delta Q_i$ are the discrepancies between the expressions on the left and right hand sides of equation \eqref{eq: act_power_balance} and \eqref{eq: react_power_balance}, respectively, and $J$ is the Jacobian given by
\begin{equation}
       J = \begin{bmatrix}
        \partial \Delta P_i/\partial \theta_i & \partial \Delta P_i/\partial |V_i|\\[2pt]
        \partial \Delta Q_i/\partial \theta_i & \partial \Delta Q_i/\partial |V_i|
       \end{bmatrix}
\end{equation}
The resulting values, $\Delta \theta_i$ and $\Delta |V_i|$, are used to update the complex bus voltages, and updates continue until an equilibrium is reached, i.e., $\Delta \theta_i = 0$ and $\Delta |V_i| = 0$. 

In practice, a single expression of the form of \eqref{eq:newton_raphson} is typically made for the full set of unknown $\theta$ and $|V|$, yielding a $(N_\text{PV} + 2N_\text{PQ})\times (N_\text{PV} + 2N_\text{PQ}) $ Jacobian matrix. To improve the computational efficiency, the problem can also be reframed in the form $A\bm{x} = \bm{b}$ to avoid computing the inverse of $J$. One can also replace $\Delta V_i$ by $\Delta V_i/V_i$ and make analogous changes in $J$ \cite{conejo2018power}. However, NR is still computationally expensive for large systems and, like most non-convex optimization procedures, the iterative update procedure is not guaranteed to converge.
The complexity of each iteration of NR is $\mathcal{O}(N^{1.4})$ and the space complexity is $\mathcal{O}(N^{1.2})$~\cite{alvarado1976computational}.

\paragraph{DCPF approximation}
In the DC power flow (DCPF) approximation the power system is linearized, yielding an approximate solution that can be found with a non-iterative and absolutely convergent procedure. To accomplish this, all reactive power flows are ignored, all voltage magnitudes are assumed to be 1 p.u., phase angle differences are assumed to be small, and branches are considered to be lossless. As a result, the two power balance equations, \eqref{eq: act_power_balance} and \eqref{eq: react_power_balance}, are reduced to the single equation~\cite{seifi2011powersystem}:
\begin{equation}
    P_i = \sum_{k=1}^{N_b} B_{ik} \theta_{ik}
\end{equation}
Thus, compared to the ACPF, a solution for $\theta$ can be found analytically, at the expense of accuracy. The time complexity of DCPF is that of a single iteration of NR, which is $\mathcal{O}(N^{1.4})$~\cite{alvarado1976computational}.  In practice, DCPF is only useful in tasks where fast computations are more important than obtaining accurate solutions~\cite{zhu2015optimization}. Nevertheless, this approximation serves as a good baseline for any model that aims to be faster than NR, while also achieving high accuracy.

\paragraph{GNN-based methods}
These methods are those that relate the most to our work. In the following, we briefly point out their limitations, which are addressed by our framework.

Boltz et al.\ propose to compute the power flow balance by applying a GNN on a pre-computed correlation matrix that determines the neighborhood of the grid buses \cite{bolz2019power}. The graph associated with such a correlation matrix does not match the physical infrastructure of the power grid and does not allow to account for the physical quantities of the power lines.

In the study from Owerko et al.~\cite{owerko2020optimal}, the top performing GNN used to solve a version of the optimal power flow problem utilizes global fully connected layers at the end, which makes the overall architecture dependent on the specific topology of the grid. This prevents the use of this GNN to account for perturbations, such as line disconnections, and to transfer the model to new grids. In general, this architecture does not fully exploit the localized computations that characterize the GNN model.
In addition, the optimal power flow problem was simplified, since the only predicted values are the active generator power outputs.

Donon et al.~\cite{donon2019graph1} propose a supervised GNN model that operates on a graph represented by multiple graph adjacency matrices, where the vertices are the transmission lines of the grid, rather than the buses. Drawbacks in their approach are the omission of branch attributes in the model, which are important to compute the power flows, and the use of a specialized architecture to implement a GNN with a large receptive field.

In a follow-up work, Donon et al.~\cite{donon2020graph2} design an unsupervised model that relies on physical equations to iteratively optimize the voltage and phase angle for each bus.
Although the experimental results show that this model can somewhat generalize to unseen topologies, the framework is quite complex and relies on a custom formulation of the power flow problem. 
Most importantly, the GNN is used to implement only a specific operation within the whole pipeline, namely to compute $\Delta \theta_i$ and $\Delta |V|_i$ from $\Delta P_i, \Delta Q_i$, which is the same type of operation performed by the NR method with a linear approximation in \eqref{eq:newton_raphson}.

Since $\Delta P_i$ and $\Delta Q_i$, the active and reactive power updates at each vertex, are computed outside the GNN using global operations that account for the whole grid topology, it is not clear how useful the local exchanges implemented by the GNN are when computing the voltage and phase angle updates.
Arguably, a simpler function approximator such as an MLP could have worked similarly well.

\section{Experiments \& Results}

The code to reproduce the experiments is publicly available online\footnote{\url{https://github.com/JonasBergHansen/Power-Flow-Balancing-with-Decentralized-Graph-Neural-Networks}}.

\subsection{Datasets and Algorithms}
\label{sec: datasets_and_algorithms}

We perform three experiments to evaluate the performance of the proposed GNN-based power flow solver to cover cases when: i) the underlying grid topology is fixed, ii) the topology is perturbed with branch disconnections, iii) multiple grids with different topologies are used to train the GNN model.

All the deep learning models are implemented in Tensorflow~\cite{abadi2016tensorflow} and we used the Spektral library~\cite{grattarola2020spektral} to build the GNN architectures.
Power grid data for training and testing is obtained by resampling values in reference case grids provided by MATPOWER \cite{zimmerman2011matpower}, an open-source Matlab library for solving steady-state power system simulation and optimization problems, such as power flow balancing. 
MATPOWER is also used to compute the solutions with the DCPF and the NR optimizer. 
Outputs of the NR optimizer are used as targets for training the deep learning models.

\paragraph{Data Acquisition}
\label{sec:acquisition}

\begin{table}[!ht]
    \caption{Grid overview}
    \centering
    \begin{tabular}{l p{5.3cm} p{5.3cm}}
        \toprule
        \textbf{Grid} & \textbf{Details}\\
        \midrule
        case9 &  9 bus, 3 generator; based on data from \cite{chow1982time}.\\
        \midrule
        case14 & Conversion of the IEEE 14 bus test case.\\
        \midrule
        case30 & Conversion of the IEEE 30 bus test case. Named case\_ieee30 in MATPOWER.\\
        \midrule
        case39 & 39 bus New England system. Modified version of data from \cite{bills1970line}.\\
        \midrule
        case57 & Conversion of the IEEE 57 bus test case.\\
        \midrule
        case89pegase & Small part of an European system stemming from the Pan European Grid Advanced Simulation and State Estimation (PEGASE) project \cite{josz2016ac, fliscounakis2013contingency}.\\
        \midrule
        case118 & Conversion of the IEEE 118 bus test case.\\
        \midrule
        case300 & Conversion of the IEEE 300 bus test case.\\
        \bottomrule
    \end{tabular}
    \label{tab:grid_info}
\end{table}
Most of the grid systems considered here are based on IEEE systems, whose overview is given in Table \ref{tab:grid_info}. 
To ensure that a feasible solution exists, only grids that are successfully solved by the NR solver in MATPOWER are included in the dataset. 
The resampling procedure we adopted is detailed in the following.

\begin{itemize}
    \item Active and reactive loads on buses are uniformly sampled between 50\% and 150\% of the reference values.
    \item Shunt susceptances and conductances are uniformly sampled between 75\% and 125\% of the reference values.
    \item Active power of generators is uniformly sampled from the reference values in a range going from 75\% up to the lowest of 125\% and the maximum allowed power output $P_{g,\text{max}}$. There is a chance that the sampled production becomes so high that the slack bus ends up with a negative active power injection. Since this occurs more often for certain grids, to avoid creating a bias in the dataset the sampling procedure is reset whenever this occurs.
    \item Voltage magnitude for generators is uniformly sampled from 0.95 p.u.\ to 1.05 p.u.
    \item Branch resistances, reactances and charging susceptances are uniformly sampled between 90\% and 110\% of the reference values.
    \item Branch transformer off nominal turns ratio is uniformly sampled from 0.8 to 1.2. The shift angle for the transformers are uniformly sampled from -0.2 rad to 0.2 rad.
\end{itemize}

\noindent While quantities such as branch impedances do not frequently change in a grid, to train the deep learning model with grid samples with the same topology but differing quantities gives it the opportunity to understand how these quantities affect the resulting power flows, thus improving the generalization across different grids.

Large spreads in the input values encourage a model to learn large weights. A model with large weights may experience instability during learning, become over-sensitive to input values, and achieve poor generalization capabilities.
Similarly, a large spread of values in the output variables may result in large error gradients that dramatically change the weight values and make the learning process unstable~\cite{bishop1995neural}.
By converting all the quantities of interest to per unit measures (p.u.), different grids will have more similar value distributions compared to using SI units.
In return, this facilitates the application of the same deep learning model to different grids.

\paragraph{Deep Learning Models}
The proposed GNN model consists of two fully-connected (FC) pre-processing layers, five ARMA$_2^8$ layers, and two post-processing layers. All of these layers have 64 hidden units, and use Leaky ReLU activations with a negative slope coefficient of $\alpha=0.2$. At the end there is a FC layer with 8 output units and a linear activation. Due to the use of 5 ARMA layers with 8 iterations each, the model implements a total of 40 propagation steps. This defines the receptive field of the GNN, i.e., the GNN allows each graph vertex to exchange information with the neighbors within 40 hops on the graph.
This configuration was found through a hyperparameter tuning session using Bayesian optimization on the data for the second experiment.
As a benchmark, we also consider a GNN model where the ARMA layers are replaced with conventional GCN layers, discussed in Section~\ref{sec:GNNs}.
Since each GCN layer implements only a single propagation step, we built a GNN with 40 GCN layers of 64 hidden units to obtain the same receptive field of the GNN with ARMA layers.

The Local MLP is designed to have roughly the same number of trainable parameters as the ARMA GNN. Specifically, the Local MLP consists of three hidden layers with 256 units, two hidden layers with 128 units, one layer with 64 units, and an output layer with 8 units. All hidden layers are equipped with a Leaky ReLU with $\alpha=0.2$.

For the Global MLP, the number of parameters depends on the size of the grid(s) it is trained on and, thus, the size of this model is not scaled to match exactly the number of parameters of the GNN. 
Since the Global MLP contains a huge amount of parameters in the input and output layer, we limited the depth of this model to prevent overfitting and to avoid having a number of parameters that is several orders of magnitude larger than in the other architectures.
Specifically, the Global MLP model consists of bus and branch pre-processing layers with 64 units each, two hidden layers with 128 units, and an output layer with number of units equal to eight times the number of branches. The processing layers and the hidden layers use Leaky ReLU activations with $\alpha=0.2$.

\begin{figure}[!ht]
    \centering
    \includegraphics[width = 0.5\textwidth]{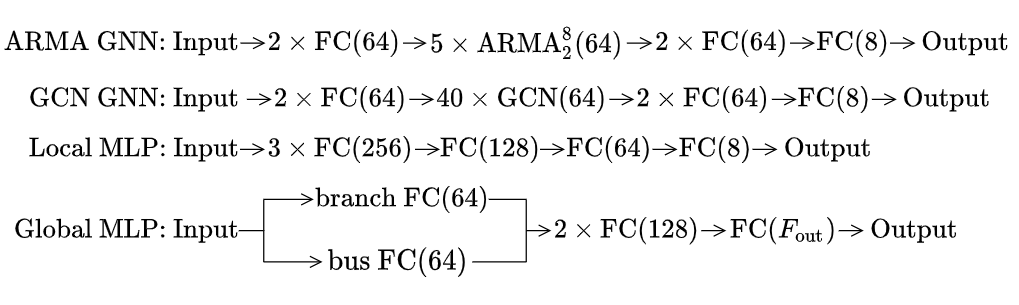}
    \caption{Overview of the architecture of the two GNN models and the two MLP models, as a combination of message passing (ARMA and GCN) and fully connected (FC) layers.
    The numbers in parentheses indicate the number of hidden/processing units. For the Global MLP model, $F_\text{out}$ is equal to eight times the number of branches of the applied grid.}
    \label{fig:final_models}
\end{figure}

An overview of the model architectures is given in Figure \ref{fig:final_models}. 
All models are trained in a supervised manner with a stochastic gradient descent learning algorithm using a mean squared error (MSE) loss between predictions and the output of the NR solver in MATPOWER.

\subsection{Constant Grid Topology}

In this first experiment the models are trained only on instances of the same grid topology, obtained with the resampling procedure described in Section~\ref{sec:acquisition}. 
This is the simplest use case, where the models only need to adapt to perturbations of the values of a specific set of components, but not to changes in the underlying graph topology.

For this task samples of three different MATPOWER case grids are considered: case30, case118 and case300. 
A total of \numprint{15000} samples were generated for each of these case grids and train/validation/test sets were arbitrarily obtained following a 56/14/30 split. 
Ten instances of the Local MLP, Global MLP, GCN and ARMA GNN models were trained for 250 epochs, using the ADAM optimizer with learning rate  \numprint{0.001} and batch size 16. After training, the weights of the model that achieved the lowest validation loss were restored. 

\begin{table}[!ht]
    \centering
    \small
    \caption{NRMSE for test set}
    \begin{tabular}{l l l l}
        \toprule
        \textbf{Method} & \textbf{case30} & \textbf{case118} & \textbf{case300}\\
        \midrule
        DCPF & $0.574$ & 0.674 & 0.745\\
        Local MLP & $0.350{ \scriptstyle \pm 0.002}$  & $0.447{ \scriptstyle \pm 0.002}$ & $0.528{ \scriptstyle \pm 0.003}$\\
        Global MLP & $0.044{ \scriptstyle \pm 0.004}$  & $0.134{ \scriptstyle \pm 0.003}$ & $0.314{ \scriptstyle \pm 0.007}$\\
        GCN & $0.611{ \scriptstyle \pm 0.03}$ & $0.887{ \scriptstyle \pm 0.02}$ & $0.851{ \scriptstyle \pm 0.02}$\\
        ARMA GNN & $\mathbf{0.022}{ \scriptstyle \pm 0.001}$ & $\mathbf{0.057}{ \scriptstyle \pm 0.002}$ & $\mathbf{0.199}{ \scriptstyle \pm 0.004}$\\
        \bottomrule
    \end{tabular}
    \label{tab:results_exp1}
\end{table}
Table \ref{tab:results_exp1} displays the normalized root mean squared error (NRMSE) values for the branch current and power predictions. The NRMSE is computed as
\begin{equation}
    \text{NRMSE}\left(\bm{Y},\hat{\bm{Y}}\right) = \frac{1}{F}\sum_{j=1}^F\sqrt{\frac{1}{N}\sum_{i=1}^N\dfrac{\left(Y_{ij} - \hat{Y}_{ij}\right)^2}{\widehat{\text{Var}}(\bm{y}_{j})}},
\end{equation}
where $F$ is the number of output features, $N$ is the number of samples and $\widehat{\text{Var}}(\bm{y}_{j})$ is the sample variance of the $j$-th output feature.
The ARMA GNN outperforms both MLP models, the GCN and the DCPF approximation. Compared to the Global MLP, the ARMA GNN achieves a performance increment ranging from roughly 40 to 60 percent. 
It should be noted that the ratio of the number of parameters between the Global MLP and the ARMA GNN is roughly 1:2 for case30, 1.5:1 for case118 and 3:1 for case300, so the difference in performance is clearly not due to a higher capacity of the GNN model.

\begin{figure*}[!ht]
    \centering
    \begin{subfigure}[b]{0.32\textwidth}
        \includegraphics[width = \textwidth]{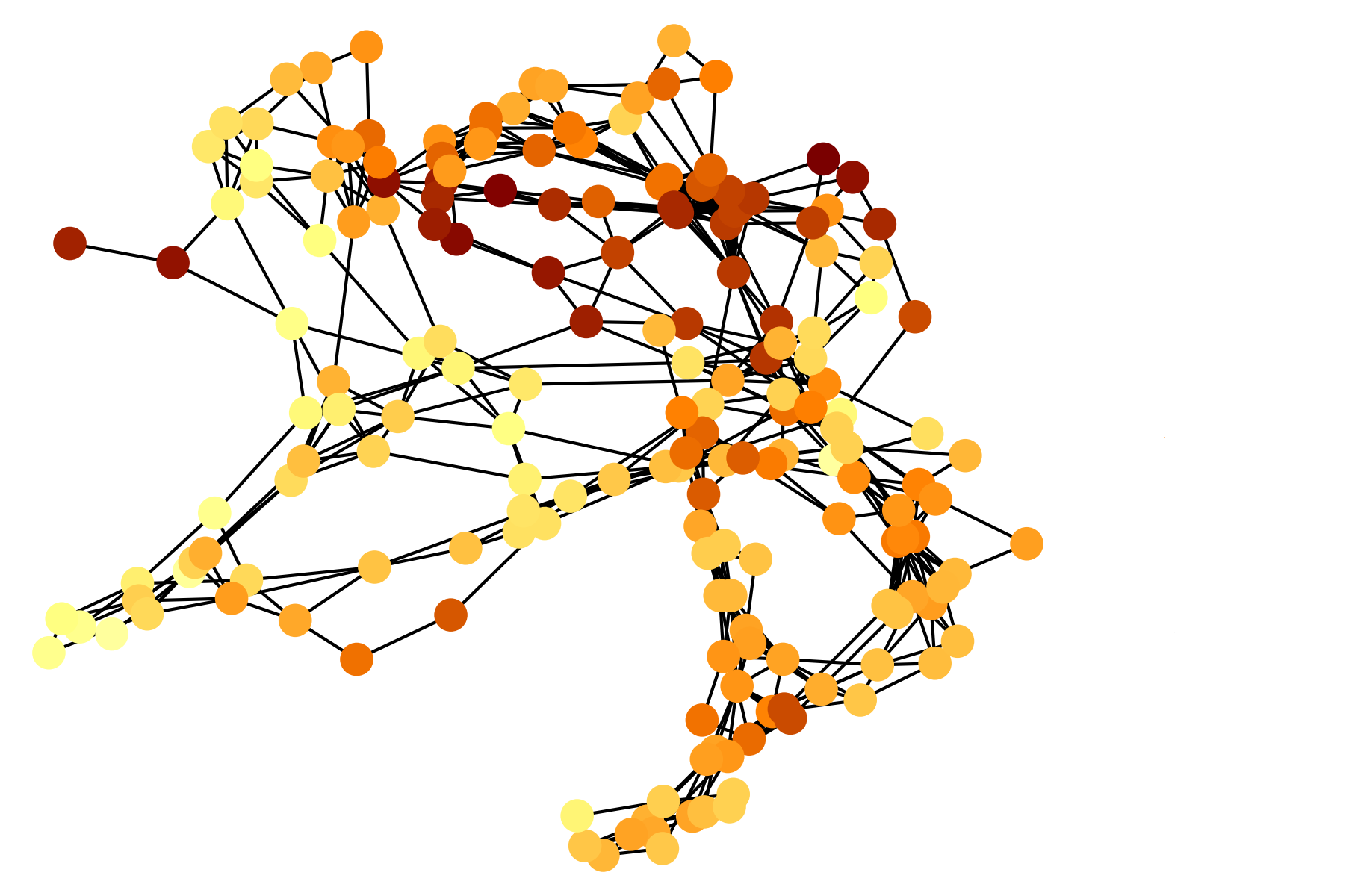}
        \caption{GCN}
    \end{subfigure}
    ~
    \begin{subfigure}[b]{0.32\textwidth}
        \includegraphics[width = \textwidth]{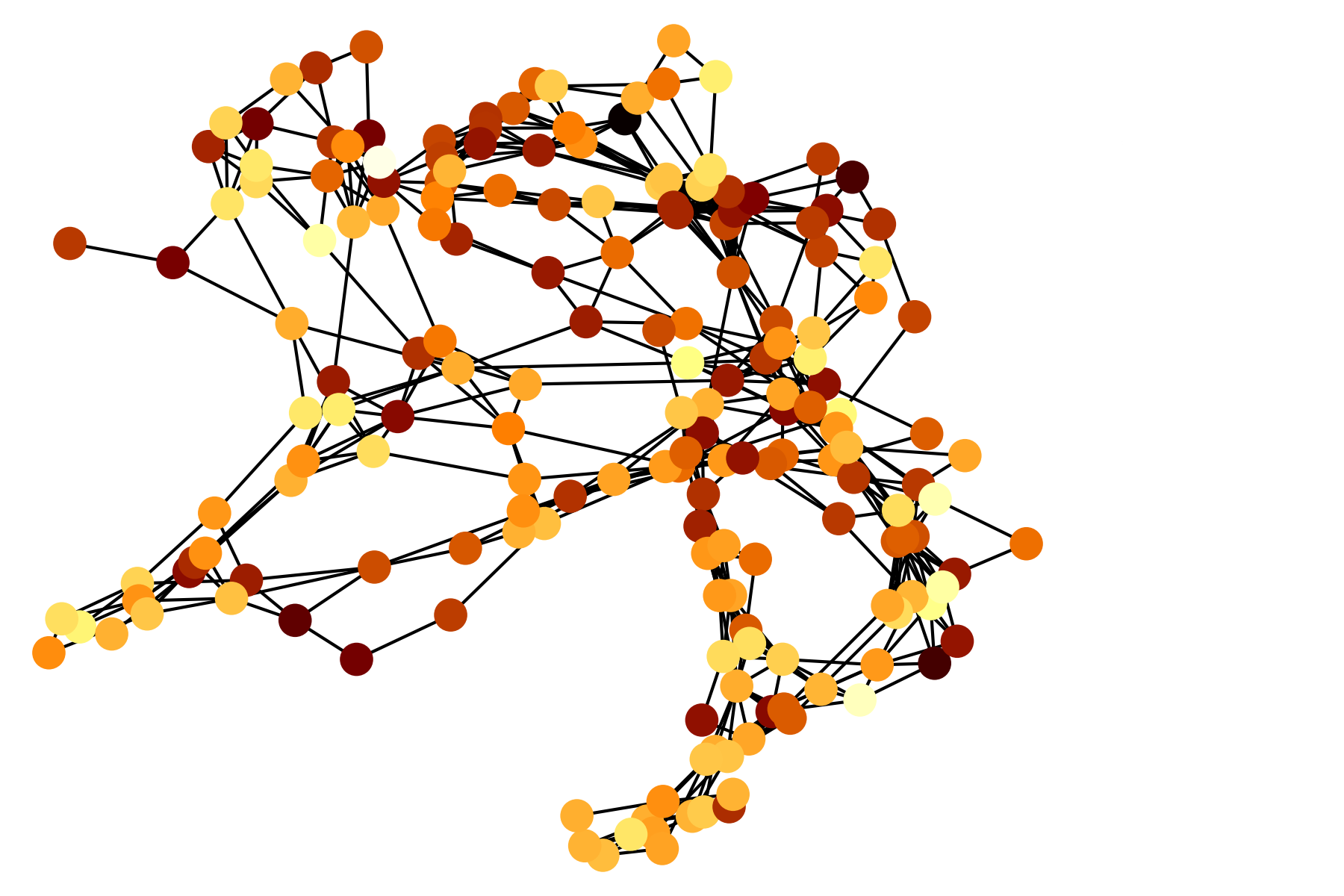}
        \caption{ARMA}
    \end{subfigure}
    ~
    \begin{subfigure}[b]{0.32\textwidth}
        \includegraphics[width = \textwidth]{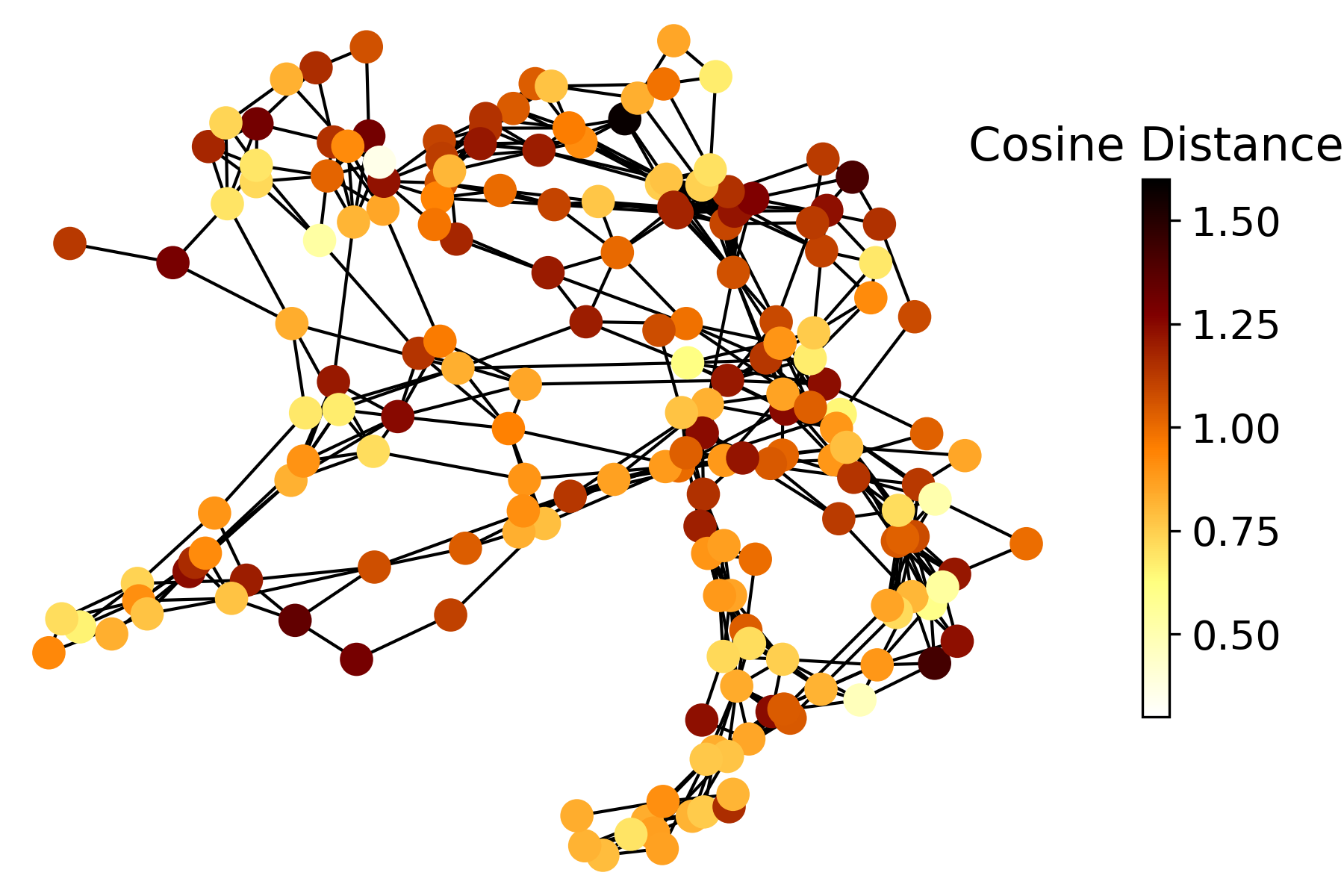}
        \caption{True values}
    \end{subfigure}
    \caption{Cosine distances between the branch values predicted by the GNNs and the mean of all the true branch values. The distances are averaged over all test samples of case118.}
    \label{fig:cosdist_case118}
 \end{figure*}

The GCN falls behind all other models in terms of NRMSE, which is likely due to the oversmoothing problem, i.e., the model is not able to produce a sharp signal in output. An illustration of the oversmoothing problem in GCN is shown in Figure \ref{fig:cosdist_case118}, which displays for each vertex of case118 the mean cosine distance between model predictions and the overall average branch label. 
The cosine distance is computed as
\begin{equation}
    \label{eq:cosine}
    \text{cosine distance} = 1 - \frac{\bm{y}_\text{pred}\cdot \overline{\bm{y}}_\text{label}}{\|\bm{y}_\text{pred}\| \|\overline{\bm{y}}_\text{label}\|}.
\end{equation}
where $\overline{\bm{y}}_\text{label}$ is the average output vector in the test set.
We obtain the cosine distances for the true labels by replacing the predictions $\bm{y}_\text{pred}$ with $\bm{y}_\text{label}$ in \eqref{eq:cosine}.
From the figure, it is clear that ARMA GNN manages to produce sharp outputs that coincide very well with the true labels, whereas the GCN only produces low-frequency signals, i.e., the output associated with neighboring vertices on the graph is very homogeneous.
Obviously, this is a strong limitation of the GCN model, since in the power flow balance state the output of neighboring vertices is not necessarily similar.
In the remaining experiments we will only consider a GNN with ARMA layers.

\begin{figure}[!ht]
    \centering
    \includegraphics[width=0.48\textwidth]{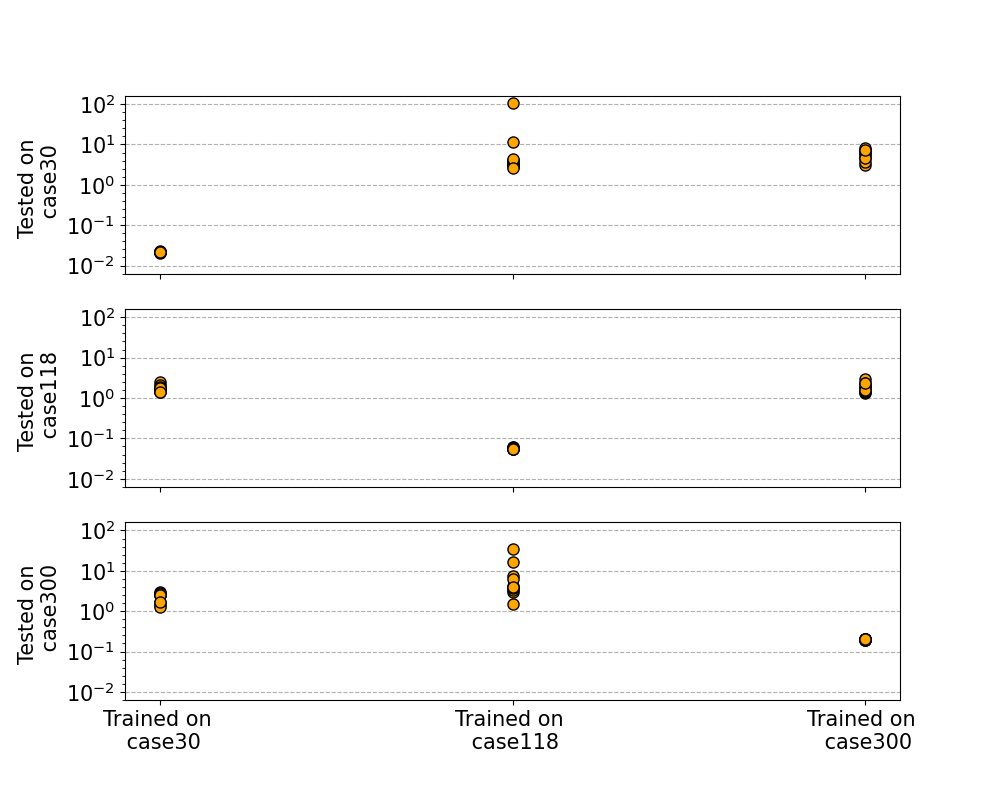}
    \caption{Scatter plot of the NRMSE scores achieved by the GNN ARMA model when trained and tested on different datasets.}
    \label{fig:cross_tests_scatter}
\end{figure}
Finally, we test how well the ARMA GNN performs when trained on samples extracted from a specific grid and then tested on samples derived from completely different power grids.
For this experiment, we consider case30, case118, and case300.
Figure \ref{fig:cross_tests_scatter} shows performance of the ARMA GNN on both the grids used for training and on the other two grids that are only used for testing.
As expected, the GNN achieves better performance when tested on samples derived from the same grid used for training.
Interestingly, the GNN trained on case118 performs particularly bad when tested on samples from case30 and case300.

\subsection{Perturbations of One Grid Topology}
In this second experiment the models are trained and tested on perturbations of a reference grid where, in addition to the sampling procedure described in Section~\ref{sec:acquisition}, parts of the network are disconnected. This setup can reflect scenarios where the grid is affected by outages or maintenance operations, and it is crucial that the power flow solver can give accurate predictions for such cases if it is meant to be used in, e.g., contingency analysis. 

To create a topology perturbation, 5-20 randomly chosen branches that are not directly connected to the slack bus are disconnected. If this causes the line graph to become disconnected, all the graph components that do not include the slack bus are removed. 
Thus, the total number of disconnections can become larger than what was initially sampled. 
With this method, there is a chance that the component with the slack bus is actually the smallest one and that most of the graph would be discarded.
To avoid this situation, any resulting grid with less than 10\% of the original buses is discarded. 
Finally, the value sampling procedure described in Section~\ref{sec:acquisition} is applied.
If the sampling fails, i.e. the NR solver does not converge, the process restarts from the disconnection step.

\begin{figure}[!ht]
    \centering
    \includegraphics[width=\columnwidth]{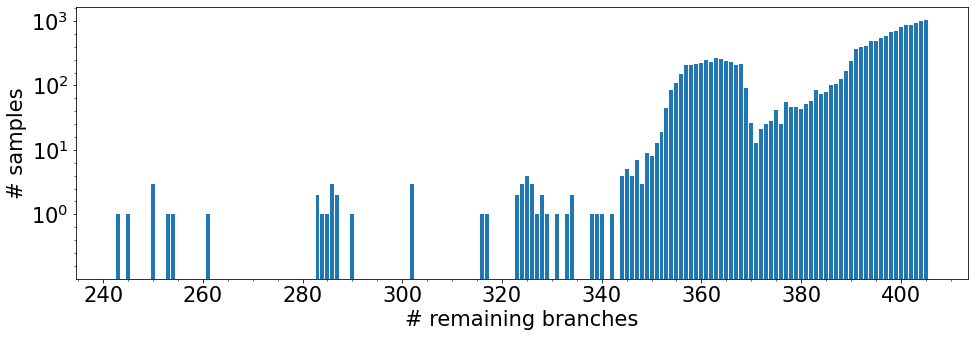}
    \caption{Bar plot showing how many perturbation samples have a given number of branches remaining (logarithmic y-axis). For reference, the full case300 grid have 411 branches.}
    \label{fig:perturbation_bar}
\end{figure}
A total of \numprint{15000} samples from case300 were generated and split into data sets following the same division as in the first experiment. Figure \ref{fig:perturbation_bar} shows the distribution of the remaining branches in the final dataset. 
Despite the initial disconnections being chosen uniformly, the resulting distribution is not uniform, since additional disconnections are triggered when a component of the graph becomes isolated. 
It is also possible that the NR solver does not converge on certain configurations, which would contribute to shape the non-uniform distribution.  

The same configuration as in the previous experiment was used to train ten instances of the Local MLP, Global MLP, and ARMA GNN models. For the Global MLP, since the parameters in the input and output layers are tied to the original topology of the case300 grid, we used a padding strategy to deal with a varying number of vertices/edges in the graph.
In particular, missing values are set to zero during inference. It is important to note that the zeros enforced here are meant to represent complete absence and not, for instance, open lines. The different perturbated grids can thus be seen as completely different grids that differ slightly in topology, but are not meant to be fully realistic contingency scenarios of a common reference grid. For an open line, resistance and reactance would be infinite, which would not be possible to set as input for the Global MLP, since it is data-driven and does not consider the physics explicitly. If these quantities are instead set to zero, the model can learn what a missing line means in practice from the relation between inputs and labels. 
However, if the grids happen to have very short lines, there would be instances where $r$ and $x$ are close to zero without being related to disconnections, in which case the MLP could struggle to distinguish between open/missing lines and very short ones (if both are present in training samples). This highlights another potential issue with MLPs in comparison to GNNs.

\begin{table}[!ht]
    \small
    \centering
    \caption{NRMSE for test set}
    \begin{tabular}{c c}
        \toprule
        \textbf{Method} & \textbf{case300 Perturbations} \\
        \midrule
        DCPF & 0.762\\ 
        Local MLP & $0.554 {\scriptstyle \pm 0.001}$ \\ 
        Global MLP & $0.371 {\scriptstyle \pm 0.004}$ \\ 
        ARMA GNN & $\textbf{0.243} {\scriptstyle \pm 0.002}$ \\ 
        \bottomrule
    \end{tabular}
    \label{tab:results_exp2}
\end{table}
The resulting NRMSE values for the test set are given in Table \ref{tab:results_exp2}. As expected, a comparison with the previous experiment shows that the performance of each model decreases in presence of disconnections. The slight increase of NRMSE for DCPF suggests that when perturbing the underlying topology, the resulting samples might, on average, violate the DCPF assumptions to a greater extent. The ranking of the performance between the models remains the same as in the previous experiment. 
The ARMA GNN is still the best performing model, followed by the Global MLP, which maintains reasonable performance even under perturbations of the underlying topology.

The good performance of the Global MLP might appear surprising at first. However, it can be explained by the fact that even if the topology changes, each bus/branch in the grid is always assigned to the same input unit of the Global MLP. 
In addition, when a vertex/edge is dropped, it is simply replaced with a zero-input.
Therefore, there is still a one-to-one correspondence between the grid components and the input units, which allows the Global MLP to factor out (to a certain extent) the underlying graph topology.

\subsection{Multiple Grids with Different Topology}

In this final experiment the models are trained simultaneously on samples of six different grids: case9, case14, case30, case39, case89pegase and case118. In addition, the models are used to compute solutions for samples of case57 and case300, which are completely unseen during training. 

To let the Global MLP process samples from grids with different topology, the input and output layer are modified such that the model can accept samples from any of the grids seen during training.
In addition, the Global MLP can only process samples from unseen grids that have a number of vertices/edges less or equal to the maximum number of vertices/edges seen during training. Each processing unit in these layers are associated with a specific attribute such as active load or branch real current injection but, unlike the configuration of the two previous experiments, these units are not associated with a specific bus or branch. Thus, for samples from the smaller grids not all the units will be utilized and the input and output arrays are zero-padded. On the other hand, since the Local MLP and ARMA GNN use only localized operations that do not depend on the number of buses and edges in the grids, their architectures are unaltered.

A total of \numprint{5000} samples are generated for each of the grids. Dataset split and training configuration remained the same as for the previous two experiments, except for the batch size that was increased to 32 due to the larger total number of samples. Ten instances of the Local MLP, Global MLP and ARMA GNN were trained on the mixed grid dataset.
Table \ref{tab:results_exp3} reports the NRMSE performance obtained in testing new samples from both the grids used for training and for the two grids (case57 and case300) that were excluded from the training set.

When testing against samples coming from the 6 grids used for training, the ARMA GNN achieves top performance on all datasets except the case9 data, where the Global MLP achieves a better score. However, for the larger grids the GNN surpasses the Global MLP by a large margin, particularly for case89pegase.
Since there is no longer a one-to-one correspondence between the input units and the grid buses, the Global MLP cannot infer the graph topology from the training samples.
Indeed, the Global MLP is unable to produce accurate solutions since the same input unit must accommodate, at the same time, very different types of buses and branches from multiple grids.
This experiment highlights the advantage of using an architecture that exploits localized operations that are independent from the underlying topology.

For the unseen grids, the NRMSE scores are substantially larger than for the grids used in training.
The main issue is that the data distributions of the training and test sets are too different from each other, which violates one of the fundamental assumption in statistics and machine learning.
Therefore, it is not surprising that the simple algorithmic solution generated by DCPF, where no model is learned, is more suitable in these cases.
It is, however, worth noticing that also for the two unseen grids case57 and case300 the ARMA GNN clearly outperforms the Global MLP. 
Also, when comparing to the first experiment we see that a GNN trained on multiple topologies improves its performance on unseen grids.
This improvement comes at the expense of a small drop in performance on the samples from grids seen during training.
Finally, we notice that the Global MLP is not able to compute predictions for case300 since the number of buses/branches are greater than in the grids used for training and the model does not have sufficient input and output units available. 

\begin{table}[t]
    \centering
    \caption{NRMSE for test set}
    \begin{tabular}{c c c c c}
        \toprule
        \textbf{Dataset} & \textbf{DCPF} & \textbf{Local MLP} & \textbf{Global MLP} & \textbf{ARMA GNN} \\
        \midrule
        case9 & 0.512 & $0.501 {\scriptstyle \pm 0.019}$ & $\textbf{0.121} {\scriptstyle \pm 0.003}$ & $0.207 {\scriptstyle \pm 0.010}$\\
        case14 & 0.559 & $0.361 {\scriptstyle \pm 0.014}$ & $0.206 {\scriptstyle \pm 0.004}$ & $\textbf{0.188} {\scriptstyle \pm 0.005}$\\
        case30 & 0.573 & $0.411 {\scriptstyle \pm 0.011}$ & $0.313 {\scriptstyle \pm 0.006}$ & $\textbf{0.212} {\scriptstyle \pm 0.007}$\\
        case39 & 0.552 & $0.550 {\scriptstyle \pm 0.002}$ & $0.249 {\scriptstyle \pm 0.001}$ & $\textbf{0.097} {\scriptstyle \pm 0.003}$\\
        case89pegase & 0.606 & $0.462 {\scriptstyle \pm 0.008}$ & $0.791 {\scriptstyle \pm 0.006}$ & $\textbf{0.071} {\scriptstyle \pm 0.001}$\\
        case118 & 0.674 & $0.456 {\scriptstyle \pm 0.002}$ & $0.672 {\scriptstyle \pm 0.015}$ & $\textbf{0.140} {\scriptstyle \pm 0.005}$\\
        \midrule
        case57 & \textbf{0.584} & $1.213 {\scriptstyle \pm 0.079}$ & $3.168 {\scriptstyle \pm 0.338}$ & $1.345 {\scriptstyle \pm 0.232}$\\
        case300 & \textbf{0.746} & $5.361 {\scriptstyle \pm 1.499}$ & N/A & $2.478 {\scriptstyle \pm 1.636}$\\
        \bottomrule
        
    \end{tabular}
    \label{tab:results_exp3}
\end{table}

\section{Discussion \& Conclusions}

We investigated the potential of a fully localized GNN trained end-to-end to balance power flows.
To enable exchange of information between distant vertices in the graph, we built a very deep GNN architecture endowed with ARMA layers, which prevents oversmoothing of the output features.
We also adopted a line graph representation to avoid treating separate categories of vertices differently, which can be difficult to handle for a fully decentralized GNN.
Importantly, the proposed representation encompasses all the relevant physical quantities of the power system and the data format can be seamlessly processed by all the most common GNN architectures.

The experimental results show that the proposed ARMA GNN model outperforms a Global MLP with a larger capacity and a global view of the graph, even when computing solutions for a fixed grid topology. 
This demonstrates the importance of using models that directly account for the graph topology of the grid and the effectiveness of the proposed line graph representation. 

Compared to an MLP, a single GNN model can be trained on multiple grid architectures and achieves good performance on samples from the same grids used in training.
However, even if the GNN outperforms the MLP, no model is particularly effective in predicting power flows in grids completely different form these seen during training.
This is not surprising since most machine learning and statistical models generalize well only when the data in the training and in the test set are similar enough and share the same properties.

We observed that the Global MLP achieves lower performance than the ARMA GNN, yet still performs reasonably under perturbations of a specific grid topology.
The reason is that there is still a one-to-one correspondence between the input nodes of the MLP and the buses in the graph, which allows to infer the underlying topology.
Moreover, the perturbations might act as a regularization factor, which forces the MLP to not rely on all components being present at the same time.
On the other hand, the MLP is completely unable to handle, at the same time, samples coming from different grid topologies.
Similarly, it would be impossible for the MLP to handle perturbations in the form of addition of buses/branches rather than disconnections.
In this case, designing a Global MLP that matches every conceivable configuration is clearly unfeasible, while a GNN with fully localized operations would have a clear advantage.


%


\ifCLASSOPTIONcaptionsoff
  \newpage
\fi



\bibliographystyle{IEEEtran}
%

\bibliography{references}

\begin{thebibliography}{10}
\providecommand{\url}[1]{#1}
\csname url@samestyle\endcsname
\providecommand{\newblock}{\relax}
\providecommand{\bibinfo}[2]{#2}
\providecommand{\BIBentrySTDinterwordspacing}{\spaceskip=0pt\relax}
\providecommand{\BIBentryALTinterwordstretchfactor}{4}
\providecommand{\BIBentryALTinterwordspacing}{\spaceskip=\fontdimen2\font plus
\BIBentryALTinterwordstretchfactor\fontdimen3\font minus
  \fontdimen4\font\relax}
\providecommand{\BIBforeignlanguage}[2]{{%
\expandafter\ifx\csname l@#1\endcsname\relax
\typeout{** WARNING: IEEEtran.bst: No hyphenation pattern has been}%
\typeout{** loaded for the language `#1'. Using the pattern for}%
\typeout{** the default language instead.}%
\else
\language=\csname l@#1\endcsname
\fi
#2}}
\providecommand{\BIBdecl}{\relax}
\BIBdecl

\bibitem{salam2020fundamentals}
M.~A. Salam, \emph{Fundamentals of Electrical Power Systems Analysis}.\hskip
  1em plus 0.5em minus 0.4em\relax Springer, 2020.

\bibitem{8634940}
S.~Dutto, G.~Masetti, S.~Chiaradonna, and F.~D. Giandomenico, ``On extending
  and comparing newton–raphson variants for solving power-flow equations,''
  \emph{IEEE Transactions on Power Systems}, vol.~34, no.~4, pp. 2577--2587,
  2019.

\bibitem{ozcanli2020deep}
A.~K. Ozcanli, F.~Yaprakdal, and M.~Baysal, ``Deep learning methods and
  applications for electrical power systems: A comprehensive review,''
  \emph{International Journal of Energy Research}, vol.~44, no.~9, pp.
  7136--7157, 2020.

\bibitem{khodayar2020deep}
M.~Khodayar, G.~Liu, J.~Wang, and M.~E. Khodayar, ``Deep learning in power
  systems research: A review,'' \emph{CSEE Journal of Power and Energy
  Systems}, vol.~7, no.~2, pp. 209--220, 2020.

\bibitem{zhao2021chapter}
Y.~Zhao and B.~Zhang, ``Deep learning in power systems,'' in \emph{Advanced
  Data Analytics for Power Systems}, A.~Tajer, S.~M. Perlaza, and H.~V. Poor,
  Eds.\hskip 1em plus 0.5em minus 0.4em\relax UK: Cambridge University Press,
  2021, ch.~3, pp. 52--73.

\bibitem{paucar2002artificial}
V.~L. Paucar and M.~J. Rider, ``Artificial neural networks for solving the
  power flow problem in electric power systems,'' \emph{Electric Power Systems
  Research}, vol.~62, no.~2, pp. 139--144, 2002.

\bibitem{donnot2018fast}
B.~Donnot, I.~Guyon, M.~Schoenauer, A.~Marot, and P.~Panciatici, ``Fast power
  system security analysis with guided dropout,'' in \emph{26th European
  Symposium on Artificial Neural Networks}, April 2018.

\bibitem{singh2021learning}
M.~Singh, V.~Kekatos, and G.~B. Giannakis, ``Learning to solve the ac-opf using
  sensitivity-informed deep neural networks,'' \emph{IEEE Transactions on Power
  Systems}, pp. 1--1, 2021.

\bibitem{zamzam2020smartgridcomm}
A.~S. Zamzam and K.~Baker, ``Learning optimal solutions for extremely fast ac
  optimal power flow,'' in \emph{2020 IEEE International Conference on
  Communications, Control, and Computing Technologies for Smart Grids
  (SmartGridComm)}, 2020, pp. 1--6.

\bibitem{donon2019graph1}
B.~Donon, B.~Donnot, I.~Guyon, and A.~Marot, ``Graph neural solver for power
  systems,'' in \emph{2019 International Joint Conference on Neural Networks
  (IJCNN)}, 2019, pp. 1--8.

\bibitem{donon2020graph2}
B.~Donon, R.~Clément, B.~Donnot, A.~Marot, I.~Guyon, and M.~Schoenauer,
  ``Neural networks for power flow: Graph neural solver,'' \emph{Electric Power
  Systems Research}, vol. 189, p. 106547, 2020.

\bibitem{owerko2020optimal}
D.~Owerko, F.~Gama, and A.~Ribeiro, ``Optimal power flow using graph neural
  networks,'' in \emph{ICASSP 2020 - 2020 IEEE International Conference on
  Acoustics, Speech and Signal Processing (ICASSP)}, 2020, pp. 5930--5934.

\bibitem{bolz2019power}
V.~Bolz, J.~Rueß, and A.~Zell, ``Power flow approximation based on graph
  convolutional networks,'' in \emph{2019 18th IEEE International Conference On
  Machine Learning And Applications (ICMLA)}, 2019, pp. 1679--1686.

\bibitem{kim2019graph}
C.~Kim, K.~Kim, P.~Balaprakash, and M.~Anitescu, ``Graph convolutional neural
  networks for optimal load shedding under line contingency,'' in \emph{2019
  IEEE Power Energy Society General Meeting (PESGM)}, 2019, pp. 1--5.

\bibitem{wang2020probabilistic}
D.~Wang, K.~Zheng, Q.~Chen, G.~Luo, and X.~Zhang, ``Probabilistic power flow
  solution with graph convolutional network,'' in \emph{2020 IEEE PES
  Innovative Smart Grid Technologies Europe (ISGT-Europe)}, 2020, pp. 650--654.

\bibitem{nauck2021predicting}
C.~Nauck, M.~Lindner, K.~Sch{\"u}rholt, H.~Zhang, P.~Schultz, J.~Kurths,
  I.~Isenhardt, and F.~Hellmann, ``Predicting dynamic stability of power grids
  using graph neural networks,'' \emph{arXiv preprint arXiv:2108.08230}, 2021.

\bibitem{bianchi2021graph}
F.~M. Bianchi, D.~Grattarola, L.~Livi, and C.~Alippi, ``Graph neural networks
  with convolutional arma filters,'' \emph{IEEE Transactions on Pattern
  Analysis and Machine Intelligence}, pp. 1--1, 2021.

\bibitem{milano2010power}
F.~Milano, \emph{Power System Modelling and Scripting}.\hskip 1em plus 0.5em
  minus 0.4em\relax Springer, 2010.

\bibitem{goodfellow2016deep}
I.~Goodfellow, Y.~Bengio, and A.~Courville, \emph{Deep learning}.\hskip 1em
  plus 0.5em minus 0.4em\relax MIT press, 2016.

\bibitem{bacciu2020gentle}
D.~Bacciu, F.~Errica, A.~Micheli, and M.~Podda, ``A gentle introduction to deep
  learning for graphs,'' \emph{Neural Networks}, vol. 129, pp. 203--221, 2020.

\bibitem{kipf2017semi}
T.~N. Kipf and M.~Welling, ``Semi-supervised classification with graph
  convolutional networks,'' in \emph{ICLR}, 2017.

\bibitem{velivckovic2018graph}
P.~Veli{\v{c}}kovi{\'c}, G.~Cucurull, A.~Casanova, A.~Romero, P.~Lio, and
  Y.~Bengio, ``Graph attention networks,'' in \emph{ICLR}, 2018.

\bibitem{hamilton2017inductive}
W.~L. Hamilton, R.~Ying, and J.~Leskovec, ``Inductive representation learning
  on large graphs,'' in \emph{NIPS}, 2017.

\bibitem{li2018deeper}
Q.~Li, Z.~Han, and X.-M. Wu, ``Deeper insights into graph convolutional
  networks for semi-supervised learning,'' in \emph{AAAI}, 2018.

\bibitem{Simonovsky_2017_CVPR}
M.~Simonovsky and N.~Komodakis, ``Dynamic edge-conditioned filters in
  convolutional neural networks on graphs,'' in \emph{Proceedings of the IEEE
  Conference on Computer Vision and Pattern Recognition (CVPR)}, July 2017.

\bibitem{harary1960some}
F.~Harary and R.~Z. Norman, ``Some properties of line digraphs,''
  \emph{Rendiconti del Circolo Matematico di Palermo}, vol.~9, no.~2, pp.
  161--168, 1960.

\bibitem{you2020design}
J.~You, Z.~Ying, and J.~Leskovec, ``Design space for graph neural networks,''
  in \emph{NeurIPS}, 2020.

\bibitem{conejo2018power}
A.~J. Conejo and L.~Baringo, \emph{Power System Operations}.\hskip 1em plus
  0.5em minus 0.4em\relax Springer, 2018.

\bibitem{alvarado1976computational}
F.~Alvarado, ``Computational complexity in power systems,'' \emph{IEEE
  Transactions on Power Apparatus and Systems}, vol.~95, no.~4, pp. 1028--1037,
  1976.

\bibitem{seifi2011powersystem}
H.~Seifi and M.~S. Sepasian, \emph{Electric Power System Planning: Issues,
  Algorithms and Solutions}.\hskip 1em plus 0.5em minus 0.4em\relax
  Springer-Verlag Berlin Heidelberg, 2011.

\bibitem{zhu2015optimization}
J.~Zhu, \emph{Optimization of Power System Operation}.\hskip 1em plus 0.5em
  minus 0.4em\relax John Wiley \& Sons, 2015.

\bibitem{abadi2016tensorflow}
\BIBentryALTinterwordspacing
M.~Abadi \emph{et~al.}, ``{TensorFlow}: Large-scale machine learning on
  heterogeneous systems,'' 2015, software available from tensorflow.org.
  [Online]. Available: \url{https://www.tensorflow.org/}
\BIBentrySTDinterwordspacing

\bibitem{grattarola2020spektral}
D.~Grattarola and C.~Alippi, ``Graph neural networks in tensorflow and keras
  with spektral [application notes],'' \emph{IEEE Computational Intelligence
  Magazine}, vol.~16, no.~1, pp. 99--106, 2021.

\bibitem{zimmerman2011matpower}
R.~D. Zimmerman, C.~E. Murillo-Sánchez, and R.~J. Thomas, ``Matpower:
  Steady-state operations, planning, and analysis tools for power systems
  research and education,'' \emph{IEEE Transactions on Power Systems}, vol.~26,
  no.~1, pp. 12--19, 2011.

\bibitem{chow1982time}
J.~H. Chow, \emph{Time-Scale Modeling of Dynamic Networks with Applications to
  Power Systems}.\hskip 1em plus 0.5em minus 0.4em\relax Springer, 1982.

\bibitem{bills1970line}
G.~Bills, ``On-line stability analysis study, rp 90-1,'' Edison Electric
  Institute, Tech. Rep., 1970.

\bibitem{josz2016ac}
C.~Josz, S.~Fliscounakis, J.~Maeght, and P.~Panciatici, ``Ac power flow data in
  matpower and qcqp format: itesla, rte snapshots, and pegase,'' \emph{arXiv
  preprint arXiv:1603.01533}, 2016.

\bibitem{fliscounakis2013contingency}
S.~Fliscounakis, P.~Panciatici, F.~Capitanescu, and L.~Wehenkel, ``Contingency
  ranking with respect to overloads in very large power systems taking into
  account uncertainty, preventive, and corrective actions,'' \emph{IEEE
  Transactions on Power Systems}, vol.~28, no.~4, pp. 4909--4917, 2013.

\bibitem{bishop1995neural}
C.~M. Bishop, \emph{Neural Networks for Pattern Recognition}.\hskip 1em plus
  0.5em minus 0.4em\relax Oxford university press, 1995.

\end{thebibliography}

%








\end{document}